\documentclass{llncs}

\usepackage{graphicx}
\usepackage{bm}

\usepackage{theorem}

\begin{document}

\title{Multi-Instance Learning by Treating Instances As Non-I.I.D. Samples}

\titlerunning{Multi-Instance Learning by Treating Instances As Non-I.I.D. Samples}

\author{Zhi-Hua Zhou \and Yu-Yin Sun \and  Yu-Feng Li}
\institute{National Key Laboratory for Novel Software Technology,\\
Nanjing University, Nanjing 210093, China\\
\{zhouzh, sunyy, liyf\}@lamda.nju.edu.cn}

\maketitle

\begin{abstract}

Previous studies on multi-instance learning typically treated instances in the \textit{bags} as
\textit{independently and identically distributed}. The instances in a bag, however, are rarely independent in
real tasks, and a better performance can be expected if the instances are treated in an non-i.i.d. way that
exploits relations among instances. In this paper, we propose two simple yet effective methods. In the first
method, we explicitly map every bag to an undirected graph and design a graph kernel for distinguishing the
positive and negative bags. In the second method, we implicitly construct graphs by deriving affinity matrices
and propose an efficient graph kernel considering the clique information. The effectiveness of the proposed
methods are validated by experiments.

\end{abstract}

\section{Introduction}

In multi-instance learning \cite{Dietterich:Lathrop:Lozano1997}, each training example is a \textit{bag} of
instances. A bag is positive if it contains at least one positive instance, and negative otherwise. Although the
labels of the training bags are known, however, the labels of the instances in the bags are unknown. The goal is
to construct a learner to classify unseen bags. Multi-instance learning has been found useful in diverse domains
such as image categorization \cite{Chen:Bi:Wang2006,Chen:Wang2004}, image retrieval \cite{Zhang:Yu:Goldman2002}
, text categorization \cite{Andrews:Tsochantaridis:Hofmann2003,Settles:Craven:Ray2008nips07}, computer security
\cite{Ruffo2000}, face detection \cite{Viola:Platt:Zhang2006,Zhang:Viola2008nips07}, computer-aided medical
diagnosis \cite{Fung:Dundar:Krishnappuram2007nips06}, etc.

A prominent advantage of multi-instance learning mainly lies in the fact that many real objects have inherent
structures, and by adopting the multi-instance representation we are able to represent such objects more
naturally and capture more information than simply using the flat single-instance representation. For example,
suppose we can partition an image into several parts. In contrast to representing the whole image as a
single-instance, if we represent each part as an instance, then the partition information is captured by the
multi-instance representation; and if the partition is meaningful (e.g., each part corresponds to a region of
saliency), the additional information captured by the multi-instance representation may be helpful to make the
learning task easier to deal with.

It is obviously not a good idea to apply multi-instance learning techniques everywhere since if the
single-instance representation is sufficient, using multi-instance representation just gilds the lily. Even on
tasks where the objects have inherent structures, we should keep in mind that the power of multi-instance
representation exists in its ability of capturing some structure information. However, as Zhou and Xu
\cite{Zhou:Xu2007} indicated, previous studies on multi-instance learning typically treated the instances in the
bags as independently and identically distributed; this neglects the fact that the relations among the instances
convey important structure information. Considering the above image task again, treating the different image
parts as inter-correlated samples is evidently more meaningful than treating them as unrelated samples.
Actually, the instances in a bag are rarely independent, and a better performance can be expected if the
instances are treated in an non-i.i.d. way that exploits the relations among instances.

In this paper, we propose two multi-instance learning methods which do not treat the instances as i.i.d. samples.
Our basic idea is to regard each bag as an entity to be processed as a whole, and regard instances as
inter-correlated components of the entity. Experiments show that our proposed methods achieve performances highly
competitive with state-of-the-art multi-instance learning methods.

The rest of this paper is organized as follows. We briefly review related work in Section 2, propose the new
methods in Section 3, report on our experiments in Section 4, conclude the paper finally in Section 5.

\section{Related Work}

Many multi-instance learning methods have been developed during the past decade.
To name a few, Diverse Density \cite{Maron:Lozano1998}, $k$-nearest neighbor algorithm Citation-$k$NN
\cite{Wang:Zucker2000}, decision trees RELIC \cite{Ruffo2000} and MITI \cite{Blockeel:Page:Srinivasan2005},
neural networks BP-MIP and RBF-MIP \cite{Zhang:Zhou2006}, rule learning algorithm RIPPER-MI
\cite{Chevaleyre:Zucker2001}, ensemble algorithms
MIBoosting \cite{Xu:Frank2004} and MILBoosting \cite{Auer:Ortner2004}, logistic regression algorithm MI-LR
\cite{Ray:Craven2005}, etc.


Kernel methods for multi-instance learning have been studied by many researchers. G\"artner et al.
\cite{Gartner:Flach:Kowalczyk2002} defined the MI-Kernel by regarding each bag as a set of feature vectors and
then applying \textit{set kernel} directly. Andrews et al. \cite{Andrews:Tsochantaridis:Hofmann2003} proposed
mi-SVM and MI-SVM. mi-SVM tries to identify a maximal margin hyperplane for the instances with subject to the
constraints that at least one instance of each positive bag locates in the positive half-space while all
instances of negative bags locate in the negative half-space; MI-SVM tries to identify a maximal margin
hyperplane for the bags by regarding margin of the ``most positive instance'' in a bag as the margin of that bag.
Cheung and Kwok \cite{Cheung:Kwok2006} argued that the sign instead of value of the margin of the most positive
instance was important. They defined a loss function which allowed bags as well as instances to participate in
the optimization process, and used the well-formed constrained concave-convex procedure to perform the
optimization. Later, Kwok and Cheung \cite{Kwok:Cheung2007} designed marginalized multi-instance kernels by
incorporating generative model into the kernel design. Chen and Wang \cite{Chen:Wang2004} proposed the DD-SVM
method which employed Diverse Density \cite{Maron:Lozano1998} to learn a set of instance prototypes and then maps
the bags to a feature space based on the instance prototypes. Zhou and Xu \cite{Zhou:Xu2007} proposed the MissSVM
method by regarding instances of negative bags as labeled examples while those of positive bags as unlabeled
examples with positive constraints. Wang et al. \cite{Wang:Yang:Zha2008} proposed the PPMM kernel by representing
each bag as some aggregate posteriors of a mixture model derived based on unlabeled data.

In addition to classification, multi-instance regression has also been studied
\cite{Amar:Dooly:Goldman2001,Ray:Page2001}, and different versions of generalized multi-instance learning have
been defined \cite{Weidmann:Frank:Pfahringer2003,Scott:Zhang:Brown2003}. The main difference between standard
multi-instance learning and generalized multi-instance learning is that in standard multi-instance learning there
is a single concept, and a bag is positive if it has an instance satisfies this concept; while in generalized
multi-instance learning \cite{Weidmann:Frank:Pfahringer2003,Scott:Zhang:Brown2003} there are multiple concepts,
and a bag is positive only when all concepts are satisfied (i.e., the bag contains instances from every concept).
Recently, research on multi-instance semi-supervised learning \cite{Rahmani:Goldman2006}, multi-instance active
learning \cite{Settles:Craven:Ray2008nips07} and multi-instance multi-label learning \cite{Zhou:Zhang2007nips06}
have also been reported. In this paper we mainly work on standard multi-instance learning
\cite{Dietterich:Lathrop:Lozano1997} and will show that our methods are also applicable to multi-instance
regression. Actually it is also possible to extend our proposal to other variants of multi-instance learning.

Zhou and Xu \cite{Zhou:Xu2007} indicated that instances in a bag should not be treated as i.i.d. samples, and
this paper provides a solution. Our basic idea is to regard every bag as an entity to be processed as a whole.
There are alternative ways to realize the idea, while in this paper we work by regarding each bag as a graph.
McGovern and Jensen \cite{McGovern:Jensen2003} have taken multi-instance learning as a tool to handle relational
data where each instance is given as a graph. Here, we are working on propositional data and there is no natural
graph. In contrast to having instances as graphs, we regard every bag as a graph and each instance as a node in
the graph.

\section{The Proposed Methods}\label{sec:migraph}

In this section we propose the MIGraph and miGraph methods. The MIGraph method explicitly maps every bag to an
undirected graph and uses a new graph kernel to distinguish the positive and negative bags. The miGraph method
implicitly constructs graphs by deriving affinity matrices and defines an efficient graph kernel considering the
clique information.

Before presenting the details, we give the formal definition of multi-instance learning as following. Let
$\mathcal{X}$ denote the instance space. Given a data set $\{(X_1, y_1),$ $\cdots, (X_i, y_i), \cdots,$ $(X_N,
y_N)\}$, where $X_i = \{\bm{x}_{i1}, \cdots, \bm{x}_{ij}, \cdots, \bm{x}_{i,n_i}\} \subseteq \mathcal{X}$ is
called a \textit{bag} and $y_i \in \mathcal{Y} = \{-1, +1\}$ is the label of $X_i$, the goal is to generate a
learner to classify unseen bags. Here $\bm{x}_{ij} \in \mathcal{X}$ is an instance $\left[ x_{ij1}, \cdots,
x_{ijl}, \cdots, x_{ijd} \right]'$, $x_{ijl}$ is the value of $\bm{x}_{ij}$ at the $l$th attribute, $N$ is the
number of training bags, $n_i$ is the number of instances in $X_i$, and $d$ is the number of attributes. If there
exists $g \in \{1,\cdots,n_i\}$ such that $\bm{x}_{ig}$ is a positive instance, then $X_i$ is a positive bag and
thus $y_i = +1$; otherwise $y_i = -1$. Yet the concrete value of the index $g$ is unknown.

\begin{figure}[!t]
\centering
\begin{minipage}[c]{1.1in}
\centering
\includegraphics[width = 1.0in]{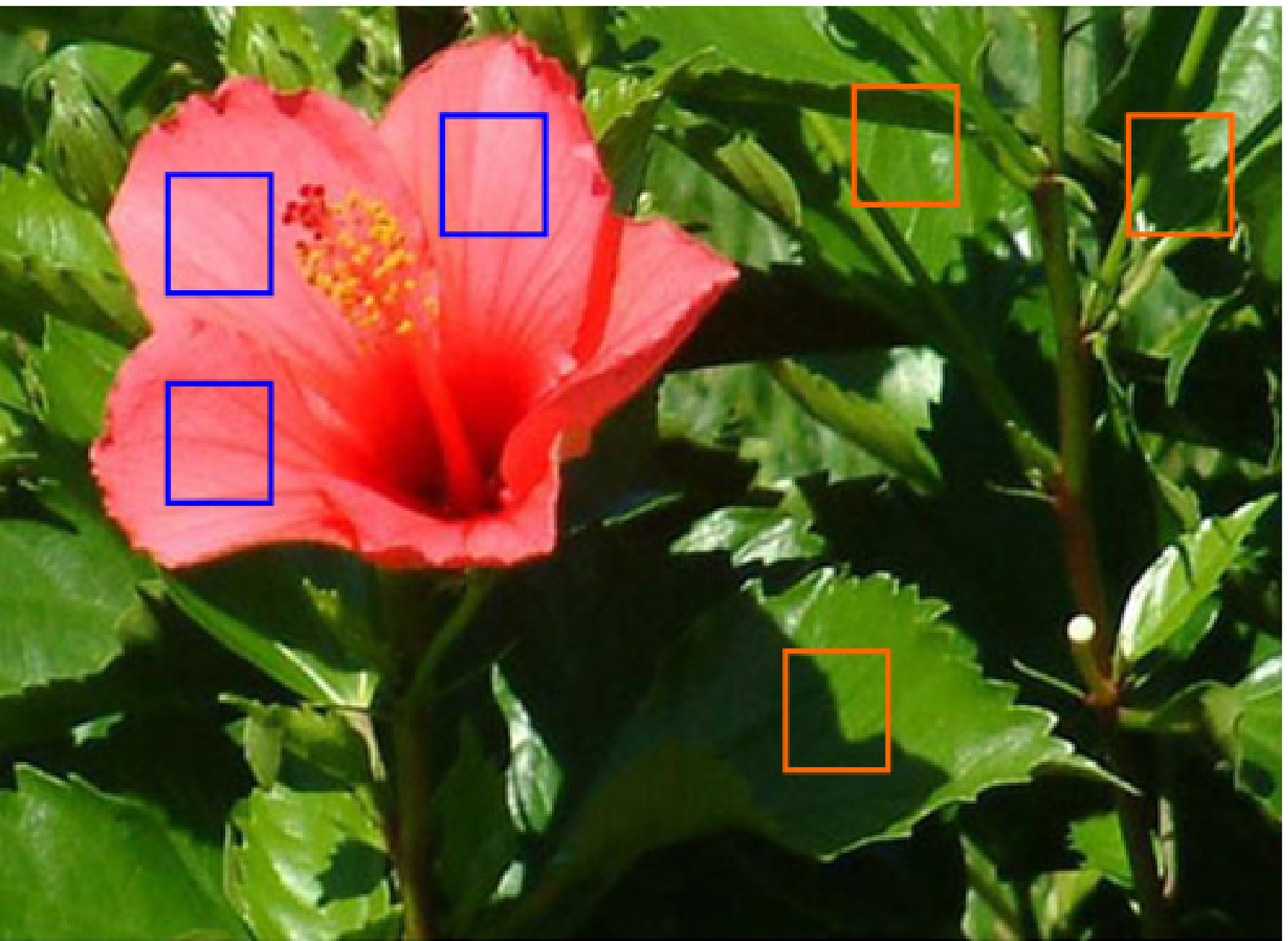}
\end{minipage}%
\begin{minipage}[c]{1.1in}
\centering
\includegraphics[width = 1.0in]{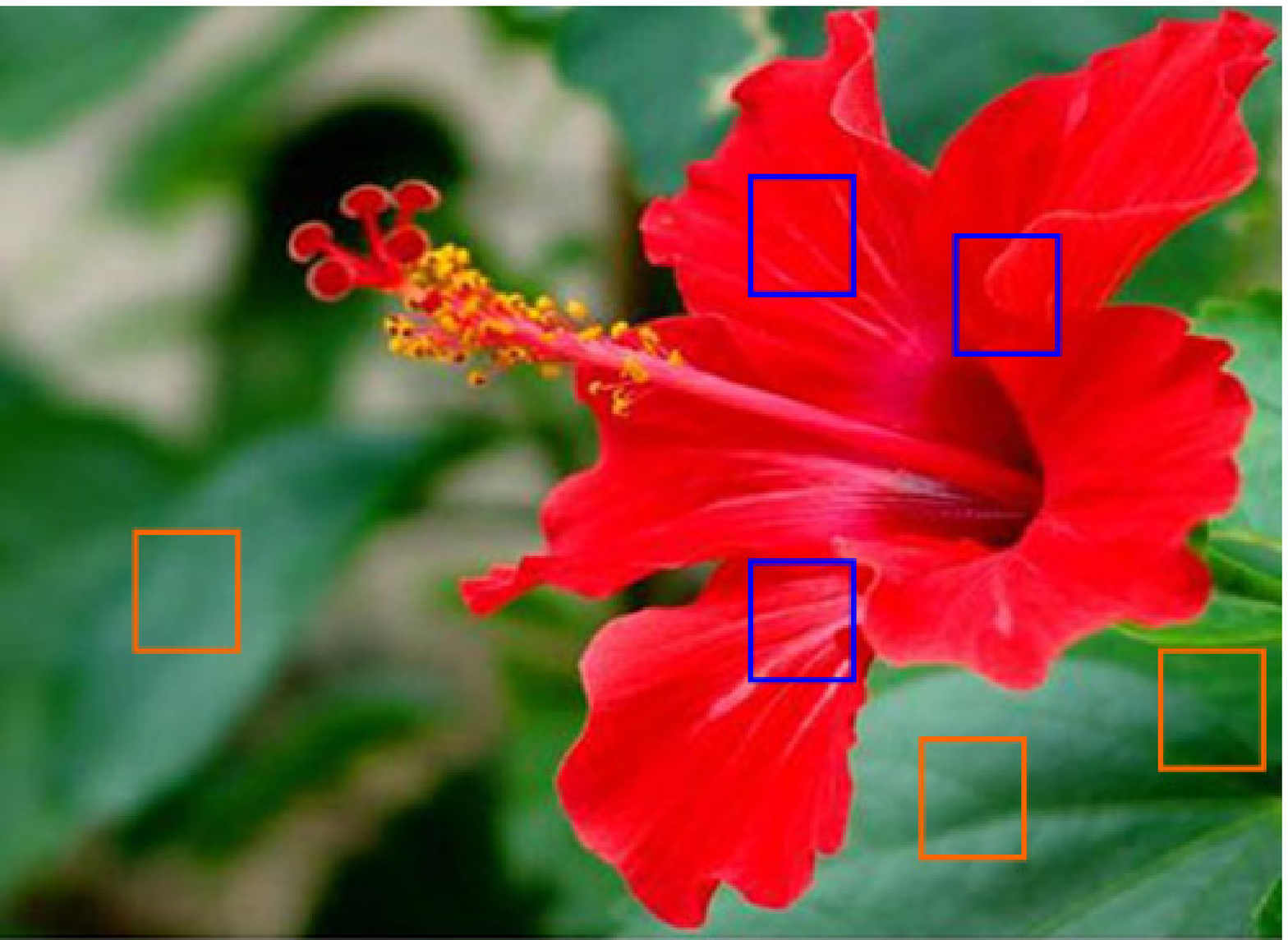}
\end{minipage}%
\begin{minipage}[c]{1.1in}
\centering
\includegraphics[width = 1.0in]{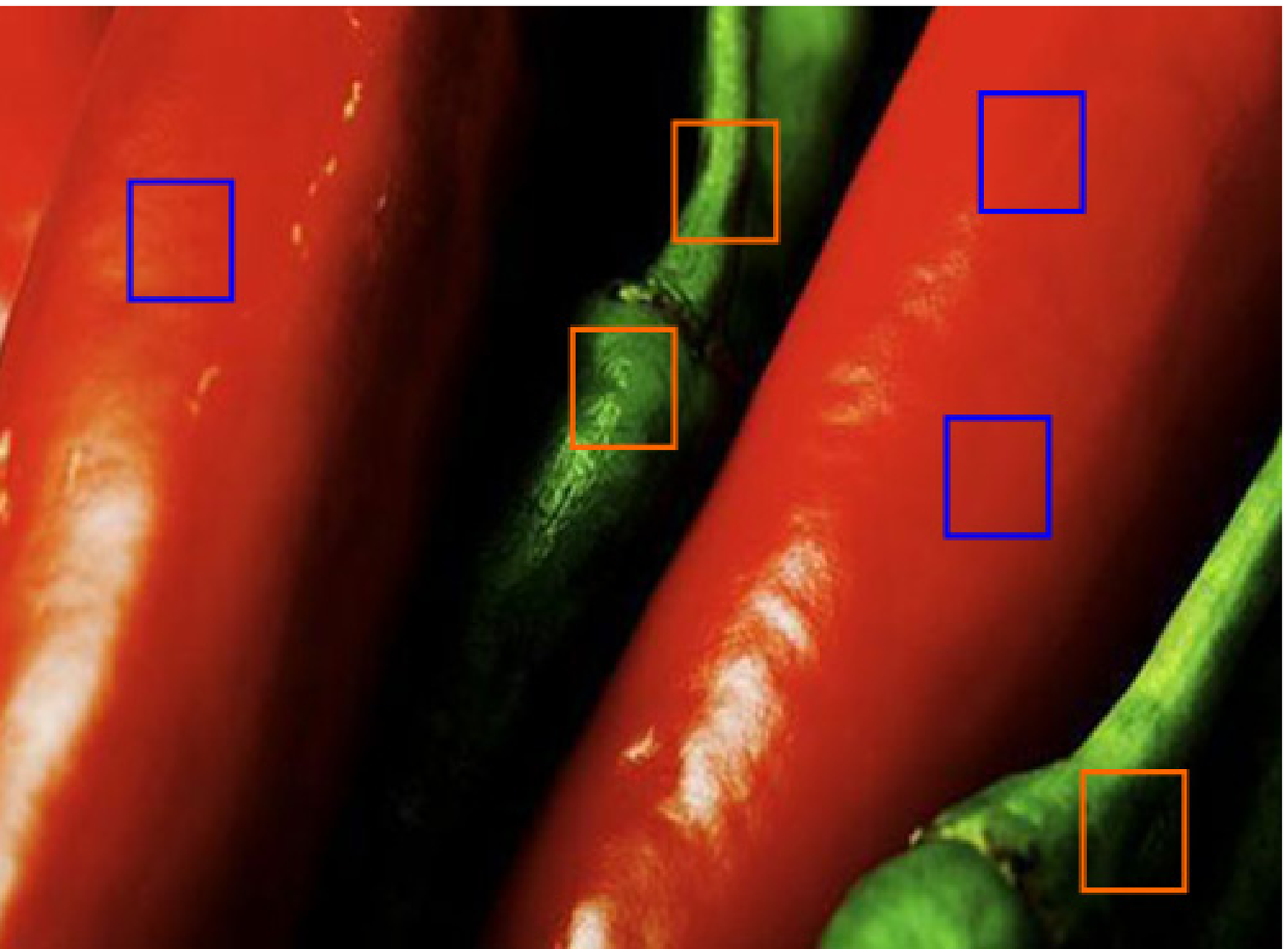}
\end{minipage}%
\caption{Example images with six marked patches each corresponding to an instance}\label{fig:rawimages}
\end{figure}

\begin{figure}[!t]
\centering
\begin{minipage}[c]{1.1in}
\centering
\includegraphics[width = 1.0in]{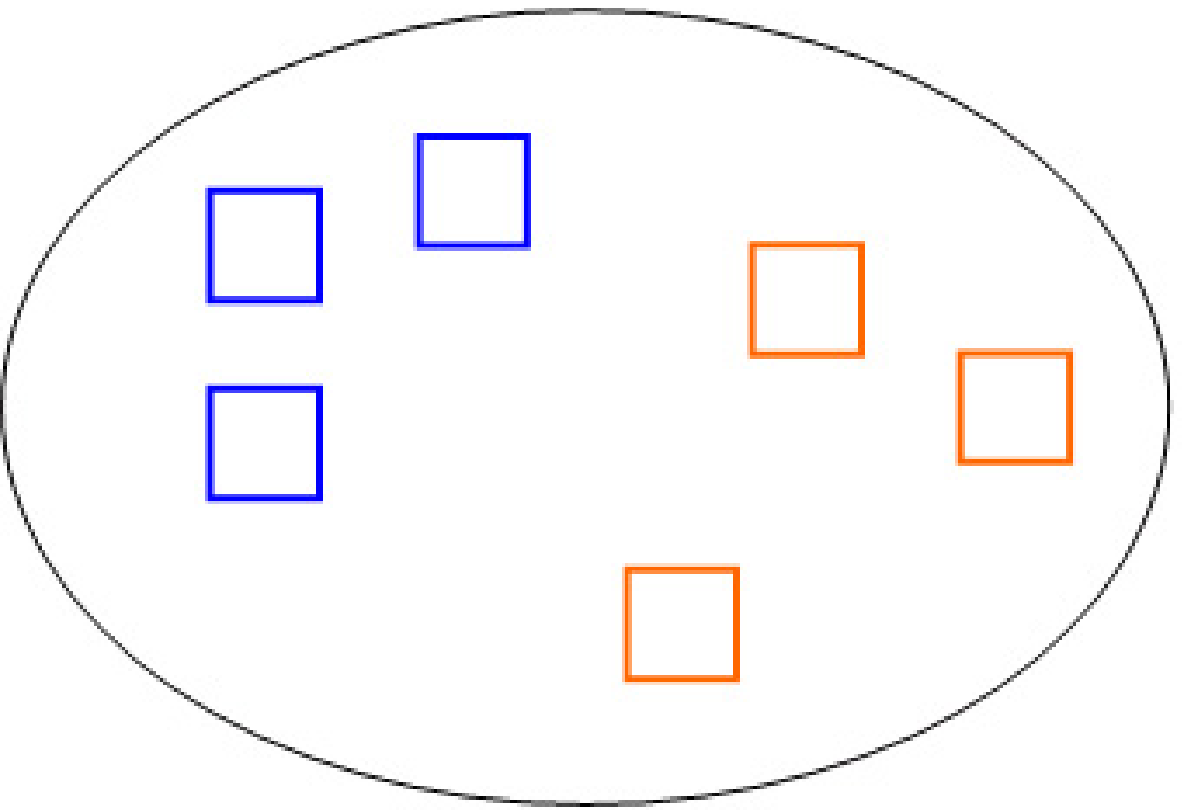}
\end{minipage}%
\begin{minipage}[c]{1.1in}
\centering
\includegraphics[width = 1.0in]{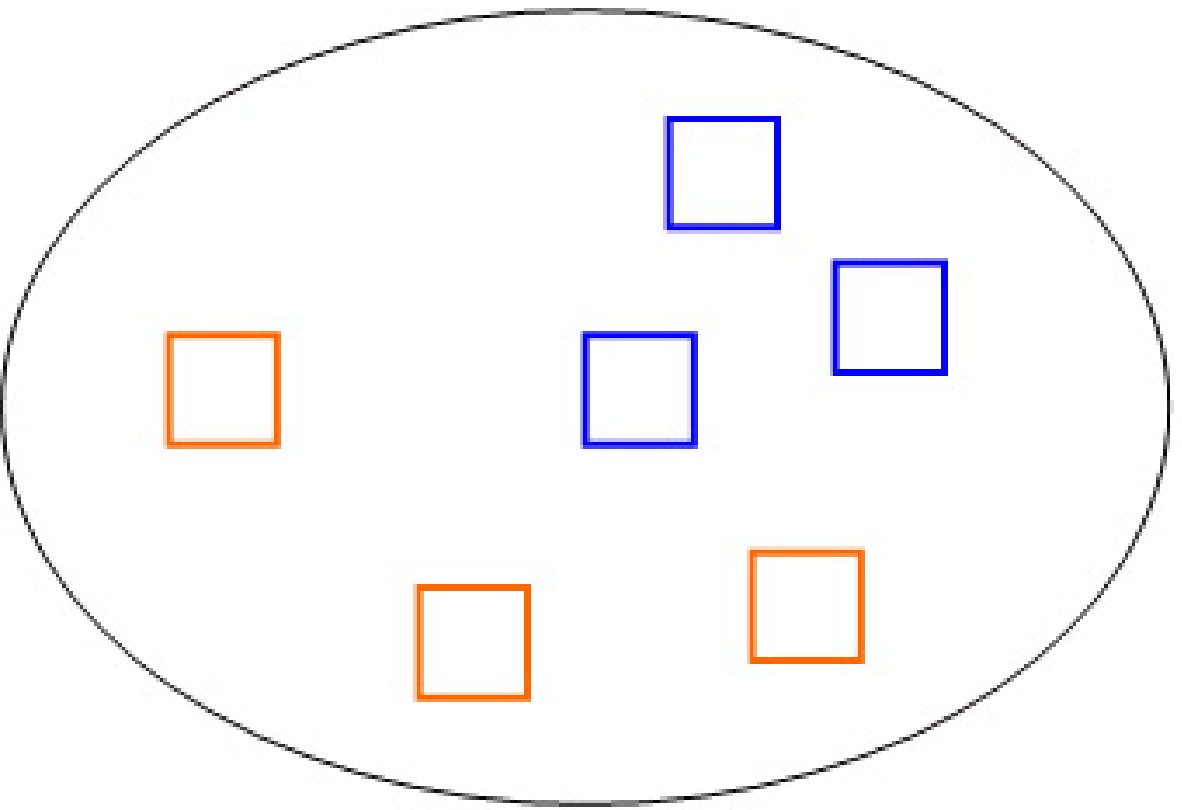}
\end{minipage}%
\begin{minipage}[c]{1.1in}
\centering
\includegraphics[width = 1.0in]{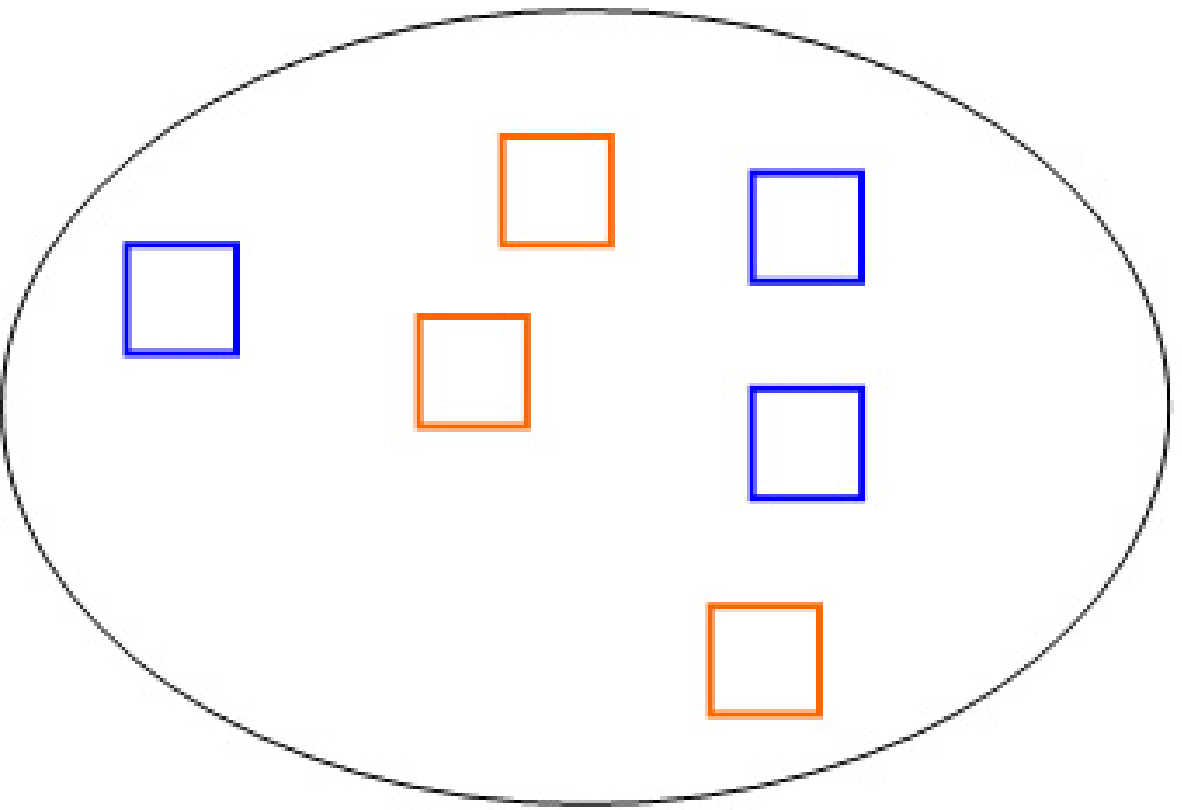}
\end{minipage}%
\caption{If we do not consider the relations among the instances, the three bags are similar to each other since
they have identical number of very similar instances}\label{fig:instanceonly}
\end{figure}

\begin{figure}[!t]
\centering
\begin{minipage}[c]{1.1in}
\centering
\includegraphics[width = 1.0in]{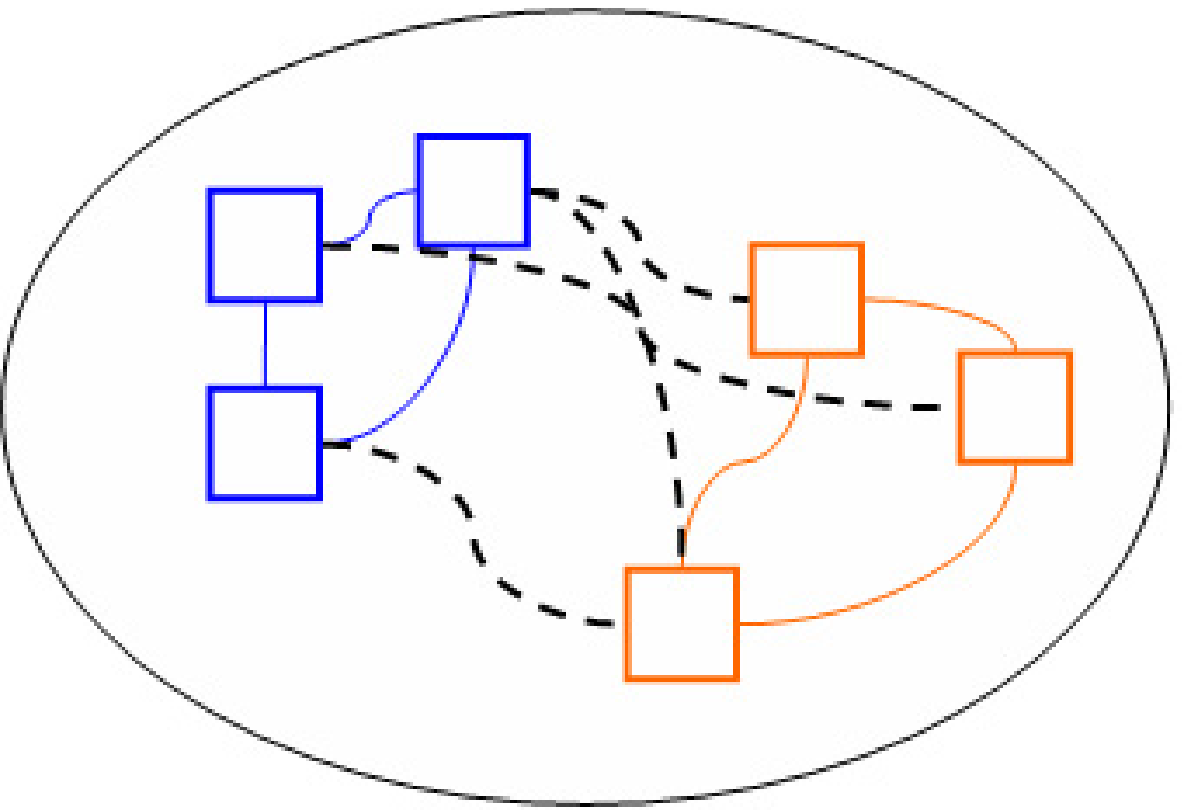}
\end{minipage}%
\begin{minipage}[c]{1.1in}
\centering
\includegraphics[width = 1.0in]{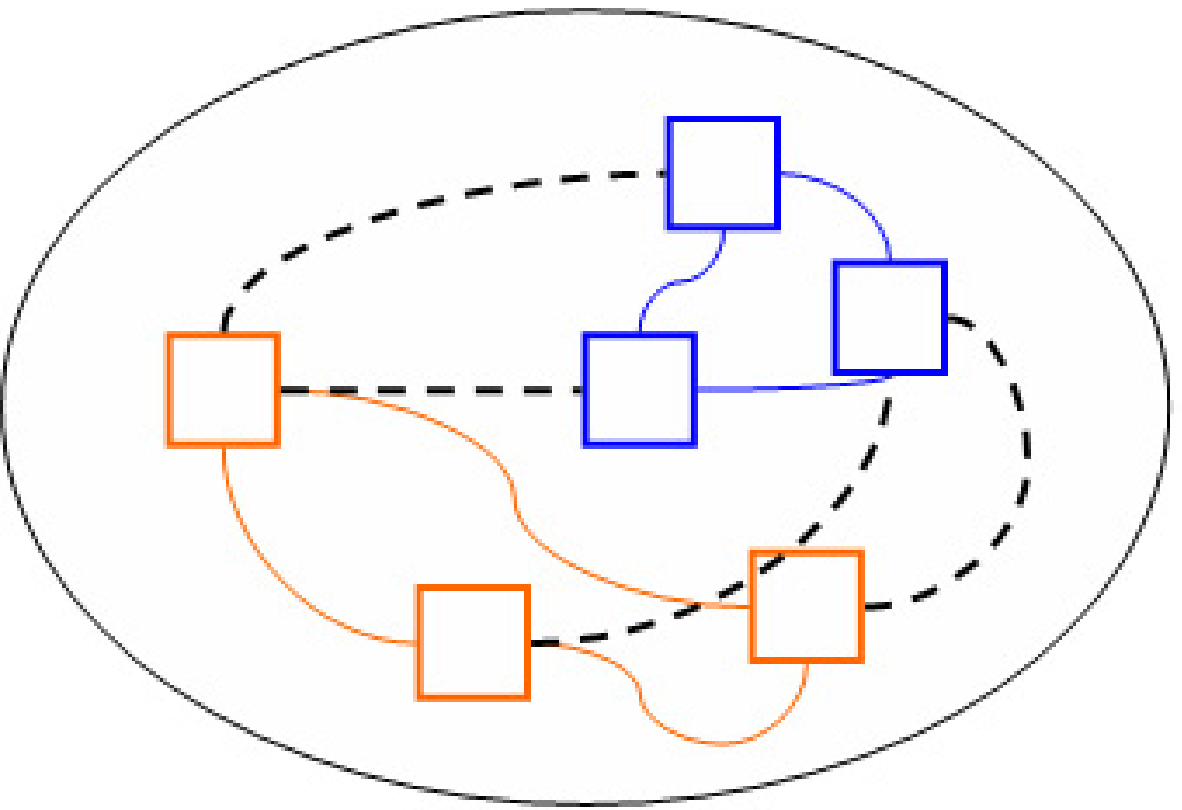}
\end{minipage}%
\begin{minipage}[c]{1.1in}
\centering
\includegraphics[width = 1.0in]{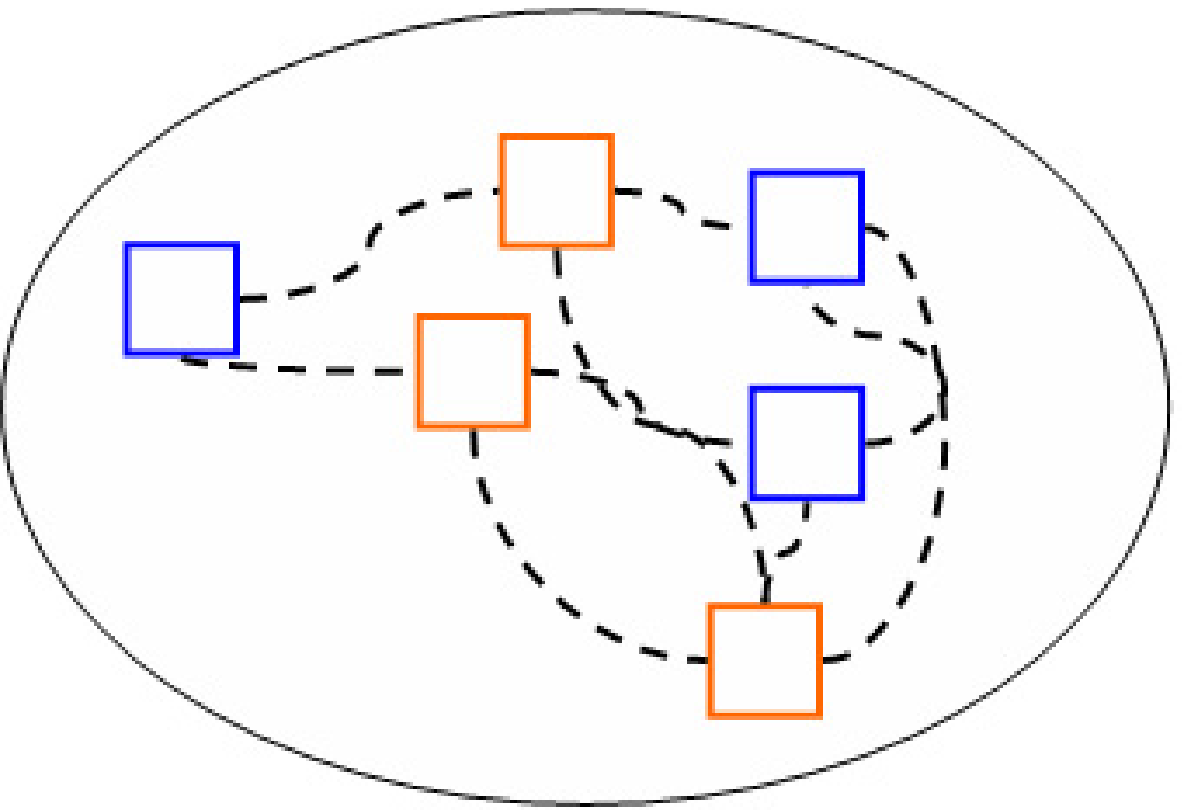}
\end{minipage}%
\caption{If we consider the relations among the instances, the first two bags are more similar than the third
bag. Here, the solid lines highlight the high affinity among similar instances}\label{fig:withrelations}
\end{figure}

We first explain our intuition of the proposed methods. Here, we use the three example images shown in
Fig.~\ref{fig:rawimages} for illustration. For simplicity, we show six marked patches in each figure, and assume
that each image corresponds to a bag, each patch corresponds to an instance in the bag, and the marked patches
with the same color are very similar (real cases are of course more complicated, but the essentials are similar
as the illustration). If the instances were treated as independent samples then Fig.~\ref{fig:rawimages} can be
abstracted as Fig.~\ref{fig:instanceonly}, which is the typical way taken by previous multi-instance learning
studies, and obviously the three bags are similar to each other since they contain identical number of very
similar instances. However, if we consider the relations among the instances, we can find that in the first two
bags the blue marks are very close to each other while in the third bag the blue marks scatter among orange
marks, and thus the first two bags should be more similar than the third bag. In this case,
Fig.~\ref{fig:rawimages} can be abstracted by Fig.~\ref{fig:withrelations}. It is evident that the abstraction in
Fig.~\ref{fig:withrelations} is more desirable than that in Fig.~\ref{fig:instanceonly}. Here the essential is
that, the relation structures of bags belonging to the same class are relatively more similar, while those
belonging to different classes are relatively more dissimilar.

Now we describe the MIGraph method. The first step is to construct a graph for each bag. Inspired by
\cite{Tenenbaum:deSilva:Langford2000} which shows that $\epsilon$-graph is helpful for discovering the underlying
manifold structure of data, here we establish an $\epsilon$-graph for every bag. The process is quite
straightforward. For a bag $X_i$, we regard every instance of it as a node. Then, we compute the distance of
every pair of nodes, e.g., $\bm{x}_{iu}$ and $\bm{x}_{iv}$. If the distance between $\bm{x}_{iu}$ and
$\bm{x}_{iv}$ is smaller than a pre-set threshold $\epsilon$, then an edge is established between these two
nodes, where the weight of the edge expresses the affinity of the two nodes (in experiments we use the normalized
reciprocal of non-zero distance as the affinity value). Many distance measures can be used to compute the
distances. According to the manifold property \cite{Tenenbaum:deSilva:Langford2000}, i.e., a small local area is
approximately an Euclidean space, we use Euclidean distance to establish the $\epsilon$-graph. When categorical
attributes are involved, we use VDM (Value Difference Metric) \cite{Stanfill:Waltz1986} as a complement. In
detail, suppose the first $j$ attributes are categorical while the remaining $(d-j)$ ones are continuous
attributes normalized to $[0, 1]$. We can use $({\sum\nolimits_{h = 1}^j {VDM({\bm x}_{1,h} ,{\bm x}_{2,h} )}} {{
+ \sum \nolimits_{h = j + 1}^d {|{\bm x}_{1,h} - {\bm x}_{2,h} |^2 } } })^{1/2}$
to measure the distance between $\bm x_1$ and $\bm x_2$. Here the VDM distance between two values $z_1$ and $z_2$
on categorical attribute $Z$ can be computed by
\begin{equation}\label{eq:vdm}
{\mathop{VDM}\nolimits} \left( {z_1,z_2} \right) = \sum\nolimits_{c = 1}^C \left| \frac{N_{Z,z_1,c}}{N_{Z,z_1}} -
\frac{N_{Z,z_2,c}}{N_{Z,z_2}} \right| ^2 \ ,
\end{equation}
where $N_{Z,z}$ denotes the number of training examples holding value $z$ on $Z$, $N_{Z,z,c}$ denotes the number
of training examples belonging to class $c$ and holding value $z$ on $Z$, and $C$ denotes the number of classes.

\begin{figure}[!t]
\centering
\includegraphics[width = 2.3in]{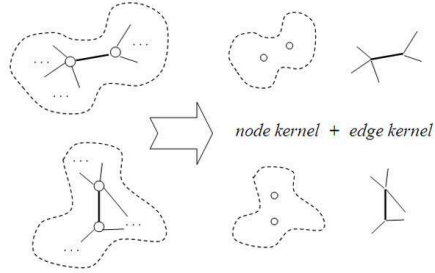}
\vspace{-2mm}\caption{Illustration of the graph kernel in MIGraph}\label{fig:idea}\vspace{-2mm}
\end{figure}

After mapping the training bags to a set of graphs, we can have many options to build a classifier. For example,
we can build a $k$-nearest neighbor classifier that employs graph edit distance \cite{Neuhaus:Bunke2007}, or we
can design a graph kernel \cite{Gartner2003} to capture the similarity among graphs and then solve classification
problems by kernel machines such as SVM. The MIGraph method takes the second way, and the idea of our graph
kernel is illustrated in Fig.~\ref{fig:idea}.

Briefly, to measure the similarity between the two bags shown in the left part of Fig.~\ref{fig:idea}, we use a
\textit{node kernel} (i.e., $k_{node}$) to incorporate the information conveyed by the nodes, use an \textit{edge
kernel} (i.e., $k_{edge}$) to incorporate the information conveyed by the edges, and aggregate them to obtain the
final graph kernel (i.e., $k_{G}$). Formally, we define $k_{G}$ as follows.

\begin{definition}
Given two multi-instance bags $X_i$ and $X_j$ which are presented as graphs $G_h(\{\bm{x}_{hu}\}_{u=1}^{n_h},$
$\{\bm{e}_{hv}\}_{v=1}^{m_h})$, $h=i,j$, where $n_h$ and $m_h$ are the number of nodes and edges in $G_h$,
respectively.
\begin{eqnarray}\label{eq:graphkernel}
k_{G}(X_i,X_j) & = & \sum\nolimits_{a=1}^{n_i}\sum\nolimits_{b=1}^{n_j} k_{node}(\bm{x}_{ia},\bm{x}_{jb}) \nonumber\\
& + & \sum\nolimits_{a=1}^{m_i}\sum\nolimits_{b=1}^{m_j} k_{edge}(\bm{e}_{ia},\bm{e}_{jb}),
\end{eqnarray}
\noindent where $k_{node}$ and $k_{edge}$ are positive semidefinite kernels. To avoid numerical problem, $k_{G}$
is normalized to
\begin{equation}\label{eq:normalization}
k_{G}(X_i,X_j) = \frac{k_{G}(X_i,X_j)}{\sqrt{k_{G}(X_i,X_i)}\sqrt{k_{G}(X_j,X_j)}} \ .
\end{equation}
\end{definition}

The $k_{node}$ and $k_{edge}$ can be defined in many ways. Here we simply define $k_{node}$ using Gaussian RBF
kernel as
\begin{equation}\label{eq:nodekernel}
k_{node} (\bm{x}_{ia},\bm{x}_{jb}) = \mbox{exp}(-\gamma ||\bm{x}_{ia} - \bm{x}_{jb}||^2),
\end{equation}
\noindent and so the first part of Eq.~\ref{eq:graphkernel} is exactly the MI-Kernel using Gaussian RBF kernel
\cite{Gartner:Flach:Kowalczyk2002}. $k_{edge}$ is also defined in a form as similar as Eq.~\ref{eq:nodekernel},
by replacing $\bm{x}_{ia}$ and $\bm{x}_{jb}$ with $\bm{e}_{ia}$ and $\bm{e}_{jb}$, respectively.

Here a key is how to define the feature vector describing an edge. In this paper, for the edge connecting the
nodes $\bm{x}_{iu}$ and $\bm{x}_{iv}$ of the bag $X_i$, we define it as $[d_u, p_u, d_v, p_v]'$, where $d_u$ is
the degree of the node $\bm{x}_{iu}$, that is, the number of edges connecting $\bm{x}_{iu}$ with other nodes.
Note that it has been normalized through dividing it by the total number of edges in the graph corresponding to
$X_i$. $d_v$ is the degree of the node $\bm{x}_{iv}$, which is defined similarly. $p_u$ is defined as $p_u =
w_{uv} / \sum w_{u,*}$, where the numerator is the weight of the edge connecting $\bm{x}_{iu}$ to $\bm{x}_{iv}$;
$w_{u,*}$ is the weight of the edge connecting $\bm{x}_{iu}$ to any nodes in $X_i$, thus the denominator is the
sum of all the weights connecting with $\bm{x}_{iu}$. It is evident that $p_u$ conveys information on how
important (or unimportant) the connection with the node $\bm{x}_{iv}$ is for the node $\bm{x}_{iu}$. $p_v$ is
defined similarly for the node $\bm{x}_{iv}$. The intuition here is that, edges are similar if properties of
their ending nodes (e.g., high-degree nodes or low-degree nodes) are similar.

The $k_{G}$ defined in Eq.~\ref{eq:graphkernel} is a positive definite kernel and it can be used for any kinds of
graphs. The computational complexity of $k_{G}(X_i,$ $X_j)$ is $O(n_in_j + m_im_j)$. The $k_{G}$ clearly
satisfies all the four major properties that should be considered for a graph kernel definition
\cite{Borgwardt:Kriegel2005}.\footnote{We have tried to apply some existing graph kernels directly but
unfortunately the results were not good. Due to the page limit, comparison with different graph kernels will be
reported in a longer version.}
Our above design is very simple, but in the next section we can see that the proposed MIGraph method is quite
effective.

A deficiency of MIGraph is that the computational complexity of $k_{G}$ is $O(n_in_j + m_im_j)$, dominated by the
number of edges. For bags containing a lot of instances, there will exist a large number of edges and MIGraph
will be hard to execute. So, it is desired to have a method with smaller computational cost. For this purpose, we
propose the miGraph method which is simple, efficient but effective.

For bag $X_i$, we can calculate the distance between its instances and derive an affinity matrix $W^i$ by
comparing the distances with a threshold $\delta$. For example, if the distance between the instances
$\bm{x}_{ia}$ and $\bm{x}_{iu}$ is smaller than $\delta$, $W^i$'s element at the $a$th row and $u$th column,
$w^{i}_{au}$, is set to 1, and 0 otherwise. There are many ways to derive $W^i$ for $X_i$. In this paper we
calculate the distances using Gaussian distance, and set $\delta$ to the average distance in the bag. The key of
miGraph, the kernel $k_g$, is defined as follows.

\begin{definition}
Given two multi-instance bags $X_i$ and $X_j$ which contains $n_i$ and $n_j$ instances, respectively.
\begin{eqnarray}\label{eq:migraph}
k_{g}(X_i,X_j) = {{\sum\nolimits_{a=1}^{n_i} \sum\nolimits_{b=1}^{n_j} {W_{ia}W_{jb} k(\bm{x}_{ia},
\bm{x}_{jb})}} \over {\sum\nolimits_{a=1}^{n_i} W_{ia} \sum\nolimits_{b=1}^{n_j} W_{jb}}},
%
%
\end{eqnarray}
\noindent where $W_{ia} = 1/ \sum\nolimits_{u=1}^{n_i} w^{i}_{au}$, $W_{jb} = 1/ \sum\nolimits_{v=1}^{n_j}
w^{j}_{bv}$, and $k(\bm{x}_{ia}, \bm{x}_{jb})$ is defined as similar as Eq.~\ref{eq:nodekernel}.
\end{definition}

To understand the intuition of $k_g$, it is helpful to consider that once we have got a good graph, instances in
one clique can be regarded as belonging to one concept. To find the cliques is generally expensive for large
graphs, while $k_g$ can be viewed as an efficient soft version of clique-based graph kernel, where the following
principles are evidently satisfied:


- 1) When $W^i = I$, i.e., every two instances do not belong to the same concept, all instances in a bag should
be treated equally, i.e., $W_{ia} = 1$ for every instance $\bm{x}_{ia}$;

- 2) When $W^i = E$ ($E$ is all-one matrix), i.e., all instances belong to the same concept, each bag can be view
as one instance and each instance contributes identically, i.e., $W_{ia} = {1 / n_i}$;

- 3) When $W^i$ is a block matrix, i.e., instances are clustered into cliques each belongs to a concept, $W_{ia}
= {1 / n_{ia}}$ where $n_{ia}$ is size of clique to which $\bm{x}_{ia}$ belongs. In this case, $k_g$ is exactly
an clique-based graph kernel;

- 4) When the value of any entries of $W^i$ increases, for example $w^i_{ab}$, $W_{ia}$ and $W_{ib}$ should
decrease since they become more similar, while other $W_{iq}$ $(q \neq a, b)$ should not be affected.



It is evident that the computational complexity of $k_g$ is as similar as that of the multi-instance kernel shown
in Eq.~\ref{eq:nodekernel}, i.e., $O(n_i n_j)$. Note that once the multi-instance kernel is obtained, the
Gaussian distances between every pair of instances have already been calculated, and it is easy to get the
$W^i$'s.


\section{Experiments}

\subsection{Benchmark Tasks}

First, we evaluate the proposed MIGraph and miGraph methods on five benchmark data sets popularly used in studies
of multi-instance learning, including \textit{Musk1}, \textit{Musk2}, \textit{Elephant}, \textit{Fox} and
\textit{Tiger}. \textit{Musk1} contains 47 positive and 45 negative bags, \textit{Musk2} contains 39 positive and
63 negative bags, each of the other three data sets contains 100 positive and 100 negative bags. More details of
the data sets can be found in \cite{Dietterich:Lathrop:Lozano1997,Andrews:Tsochantaridis:Hofmann2003}.

We compare MIGraph, miGraph with MI-Kernel \cite{Gartner:Flach:Kowalczyk2002} via ten times 10-fold cross
validation (i.e., we repeat 10-fold cross validation for ten times with different random data partitions). All
these methods use Gaussian RBF Kernel and the parameters are determined through cross validation on training
sets. The average test accuracy and standard deviations are shown in
Table~\ref{table:benchmarkresult}\footnote{We have re-implemented MI-Kernel since the comparison with MI-Kernel
will clearly show whether it is helpful to treat instances as non-i.i.d. samples (this is the only difference
between our methods and MI-Kernel). Note that the performance of MI-Kernel in our implementation is better than
that reported in \cite{Gartner:Flach:Kowalczyk2002}.}. The table also shows the performance of several other
multi-instance kernel methods, including MI-SVM and mi-SVM \cite{Andrews:Tsochantaridis:Hofmann2003}, MissSVM
\cite{Zhou:Xu2007} and PPMM kernel \cite{Wang:Yang:Zha2008}, and the famous Diverse Density algorithm
\cite{Maron:Lozano1998} and its improvement EM-DD \cite{Zhang:Goldman2002}. The results of all methods except
Diverse Density were obtained via ten times 10-fold cross validation; they were the best results reported in
literature and since they were obtained in different studies and the standard deviations were not available,
these results are only for reference instead of a rigorous comparison. The best performance on each data set is
bolded.

Table~\ref{table:benchmarkresult} shows that the performance of MIGraph and miGraph are quite good. On
\textit{Musk1} they are only worse than PPMM kernel; note that the results of PPMM kernel were obtained through
an exhaustive search that may be prohibitive in practice \cite{Wang:Yang:Zha2008}. On \textit{Musk2},
\textit{Elephant} and \textit{Fox} miGraph and MIGraph are respectively the best and second-best algorithms.
Pairwise $t$-tests at 95\% significance level indicate that miGraph is significantly better than MI-Kernel on all
data sets except that on \textit{Musk2} there is no significant difference.

\begin{table}[!t]
\begin{center}
\vspace{-2mm}\caption{Accuracy (\%) on benchmark tasks}\label{table:benchmarkresult} \small \smallskip
\begin{tabular}{lccccc}\hline\noalign{\smallskip}
  Algorithm & \textit{Musk1} & \textit{Musk2} & \textit{Elept} & \textit{Fox} & \textit{Tiger}\\\hline\noalign{\smallskip}
  MIGraph    & 90.0        & 90.0           & $85.1$      & $61.2$       & $81.9$\\
             & $\pm 3.8$   & $\pm 2.7$      & $\pm 2.8$   & $\pm 1.7$    & $\pm 1.5$\\\hline\noalign{\smallskip}
  miGraph    & $88.9$      & $\bf{90.3}$    & $\bf{86.8}$ & $\bf{61.6}$  & $\bf{86.0}$\\
             & $\pm 3.3$   & $\bf{\pm 2.6}$ & $\bf{\pm 0.7}$  & $\bf{\pm 2.8}$  & $\bf{\pm 1.6}$\\\hline\noalign{\smallskip}
  MI-Kernel   & $88.0$      & 89.3           & $84.3$      & $60.3$       & $84.2$\\
             & $\pm 3.1$   & $\pm 1.5$      & $\pm 1.6$  & $\pm 1.9$  & $\pm 1.0$\\\hline\noalign{\smallskip}
  MI-SVM     & 77.9 & 84.3 & 81.4 & 59.4 & 84.0\\
  mi-SVM     & 87.4 & 83.6 & 82.0 & 58.2 & 78.9\\
  MissSVM    & 87.6 & 80.0 & N/A  & N/A  & N/A\\
  PPMM       & $\bf{95.6}$ & 81.2 & 82.4 & 60.3 & 82.4\\
  DD         & 88.0 & 84.0 & N/A  & N/A  & N/A\\
  EM-DD      & 84.8 & 84.9 & 78.3 & 56.1 & 72.1\\
\noalign{\smallskip}\hline
\end{tabular}\vspace{-4mm}
\end{center}
\end{table}

%

\subsection{Image Categorization}

Image categorization is one of the most successful applications of multi-instance learning. The data sets
\textit{1000-Image} and \textit{2000-Image} contain ten and twenty categories of COREL images, respectively,
where each category has 100 images. Each image is regarded as a bag, and the ROIs (Region of Interests) in the
image are regarded as instances described by nine features. More details of these data sets can be found in
\cite{Chen:Wang2004,Chen:Bi:Wang2006}.

We use the same experimental routine as that described in \cite{Chen:Bi:Wang2006}. On each data set, we randomly
partition the images within each category in half, and use one subset for training while the other for testing.
The experiment is repeated for five times with five random splits, and the average results are recorded.
One-against-one strategy is used by MIGraph, miGraph and MI-Kernel for this multi-class task. Following the style
of \cite{Chen:Wang2004,Chen:Bi:Wang2006}, we present the overall accuracy as well as 95\% confidence intervals in
Table~\ref{table:CORELresult}. For reference, the table also shows the best results of some other multi-instance
learning methods reported in literature, including MI-SVM
\cite{Andrews:Tsochantaridis:Hofmann2003,Chen:Wang2004}, DD-SVM \cite{Chen:Wang2004}, $k$means-SVM
\cite{Csurka:Bray:Dance:Fan2004}, MissSVM \cite{Zhou:Xu2007} and MILES \cite{Chen:Bi:Wang2006}.

\begin{table}[!t]
\begin{center}
\vspace{-2mm}\caption{Accuracy (\%) on image categorization}\label{table:CORELresult}\vspace{-2mm}\small
\begin{tabular}{lcc}\hline
Algorithm & \textit{1000-Image} & \textit{2000-Image}\\
\hline\noalign{\smallskip}
MIGraph      & $\bf{83.9 : [81.2, 85.7]}$ & $\bf{72.1 : [71.0, 73.2]}$\\
miGraph      & $82.4 : [80.2, 82.6]$ & $70.5 : [68.7, 72.3]$\\
MI-Kernel    & $81.8 : [80.1, 83.6]$ & $72.0 : [71.2, 72.8]$\\\hline\noalign{\smallskip}
MI-SVM       & $74.7 : [74.1, 75.3]$ & $54.6 : [53.1, 56.1]$\\
DD-SVM       & $81.5 : [78.5, 84.5]$ & $67.5 : [66.1, 68.9]$\\
MissSVM      & $78.0 : [75.8, 80.2]$ & $65.2 : [62.0, 68.3]$\\
$k$means-SVM & $69.8 : [67.9, 71.7]$ & $52.3 : [51.6, 52.9]$\\
MILES        & $82.6 : [81.4, 83.7]$ & $68.7 : [67.3, 70.1]$\\
\noalign{\smallskip}\hline
\end{tabular}\vspace{-3mm}
\end{center}
\end{table}

It can be found from Table~\ref{table:CORELresult} that on the image categorization task our proposed MIGraph and
miGraph are highly competitive with state-of-the-art multi-instance learning methods. In particular, MIGraph is
the best performed method. This confirms our intuition that MIGraph is a good choice when each bag contains a few
instances, and miGraph is better when each bag contains a lot of instances.

By examining the detail results on \textit{1000-Image}, we found that both MIGraph and miGraph or at least one of
them are better than MI-Kernel on most categories, except on \textit{African} and \textit{Dinosaurs}. This might
owe to the fact that the structure information of examples belonging to these complicated
concepts\footnote{\textit{Dinosaurs} is complicated since it contains many different kinds of imaginary animals,
toys and even bones.} is too difficult to be captured by the simple schemes used in MIGraph and miGraph, while
using incorrect structure information is worse than conservatively treating the instances as i.i.d. samples.

For all three methods the largest errors occur between \textit{Beach} and \textit{Mountains} (the full name of
this category is \textit{Mountains \& glaciers}). This phenomenon has been observed before
\cite{Chen:Wang2004,Chen:Bi:Wang2006,Zhou:Xu2007}, owing to the fact that many images of these two categories
contain semantically related and visually similar regions such as those corresponding to mountain, river, lake
and ocean.

\subsection{Text Categorization}

The twenty text categorization data sets were derived from the \textit{20 Newsgroups} corpus
popularly used in text categorization. Fifty positive and fifty negative bags were generated for each of the 20
news categories. Each positive bag contains 3\% posts randomly drawn from the target category and the other
instances (and all instances in negative bags) randomly and uniformly drawn from other categories. Each instance
is a post represented by the top 200 TFIDF features.

On each data set we run ten times 10-fold cross validation (i.e., we repeat 10-fold cross validation for ten
times with different random data partitions). MIGraph does not return results in a reasonable time, and so we
only present the average accuracy with standard deviations of miGraph and MI-Kernel in
Table~\ref{table:textresult}, where the best result on each data set is bolded.

\begin{table}[!t]
\begin{center}
\vspace{-2mm}\caption{Accuracy (\%) on text categorization}\label{table:textresult}\smallskip\small
\begin{tabular}{lcc}\hline\noalign{\smallskip}
Data set & MI-Kernel & miGraph\\
\hline\noalign{\smallskip}
\textit{alt.atheism}    & $60.2 \pm 3.9$ & $\bf{65.5 \pm 4.0}$\\
\textit{comp.graphics}  & $47.0 \pm 3.3$ & $\bf{77.8 \pm 1.6}$\\
\textit{comp.os.ms-windows.misc}  & $51.0 \pm 5.2$ & $\bf{63.1 \pm 1.5}$\\
\textit{comp.sys.ibm.pc.hardware} & $46.9 \pm 3.6$ & $\bf{59.5 \pm 2.7}$\\
\textit{comp.sys.mac.harware}     & $44.5 \pm 3.2$ & $\bf{61.7 \pm 4.8}$\\
\textit{comp.window.x}  & $50.8 \pm 4.3$ & $\bf{69.8 \pm 2.1}$\\
\textit{misc.forsale}   & $51.8 \pm 2.5$ & $\bf{55.2 \pm 2.7}$\\
\textit{rec.autos}      & $52.9 \pm 3.3$ & $\bf{72.0 \pm 3.7}$\\
\textit{rec.motorcycles}     & $50.6 \pm 3.5$ & $\bf{64.0 \pm 2.8}$\\
\textit{rec.sport.baseball}  & $51.7 \pm 2.8$ & $\bf{64.7 \pm 3.1}$\\
\textit{rec.sport.hockey}    & $51.3 \pm 3.4$ & $\bf{85.0 \pm 2.5}$\\
\textit{sci.crypt}        & $56.3 \pm 3.6$ & $\bf{69.6 \pm 2.1}$\\
\textit{sci.electronics}  & $50.6 \pm 2.0$ & $\bf{87.1 \pm 1.7}$\\
\textit{sci.med}          & $50.6 \pm 1.9$ & $\bf{62.1 \pm 3.9}$\\
\textit{sci.space}        & $54.7 \pm 2.5$ & $\bf{75.7 \pm 3.4}$\\
\textit{sci.religion.christian}  & $49.2 \pm 3.4$ & $\bf{59.0 \pm 4.7}$\\
\textit{talk.politics.guns}      & $47.7 \pm 3.8$ & $\bf{58.5 \pm 6.0}$\\
\textit{talk.politics.mideast}   & $55.9 \pm 2.8$ & $\bf{73.6 \pm 2.6}$\\
\textit{talk.politics.misc}      & $51.5 \pm 3.7$ & $\bf{70.4 \pm 3.6}$\\
\textit{talk.religion.misc}      & $55.4 \pm 4.3$ & $\bf{63.3 \pm 3.5}$\\
\noalign{\smallskip}\hline
\end{tabular}
\end{center}\vspace{-4mm}
\end{table}

Pairwise $t$-tests at 95\% significance level indicate that, miGraph is significantly better than MI-Kernel on
all the text categorization data sets. It is impressive that, by examining the detail results we found that if we
consider 
each time of the ten times 10-fold cross validation, the number of win/tie/lose of miGraph versus MI-Kernel is
10/0/0 on 16 out of the 20 data sets, 9/0/1 on two data sets (\textit{talk.politics.guns} and
\textit{talk.religion.misc}), and 7/2/1 on the other two data sets (\textit{alt.atheism} and
\textit{misc.forsale}).

\subsection{Multi-Instance Regression}

We also compare MIGraph, miGraph and MI-Kernel on four multi-instance regression data sets, including
LJ-160.166.1, LJ-160.166.1-S, LJ-80.166.1 and LJ-80.166.1-S (abbreviated as LJ160.1, LJ160.1S, LJ80.1 and
LJ80.1S, respectively). In the name LJ-\textit{r.f.s}, $r$ is the number of relevant features, $f$ is the number
of features, and $s$ is the number of \textit{scale factors} used for the relevant features that indicate the
importance of the features. The suffix $S$ indicates that the data set uses only labels that are not near 1/2.
More details of these data sets can be found in \cite{Amar:Dooly:Goldman2001}.

We perform leave-one-out tests and report the results
in Table~\ref{table:Regressionresult}. For reference, the table also shows the leave-one-out results of some
other methods reported in literature, including Diverse Density \cite{Maron:Lozano1998,Amar:Dooly:Goldman2001},
BP-MIP and RBF-MIP \cite{Zhang:Zhou2006}. In Table~\ref{table:Regressionresult} the best performance on each data
set is bolded. It is evident that our proposed miGraph and MIGraph methods also work well on multi-instance
regression tasks.

\begin{table}[!t]
\begin{center}
\vspace{-2mm}\caption{Squared loss on multi-instance regression
tasks}\label{table:Regressionresult}\smallskip\small
\begin{tabular}{lcccc}\hline\noalign{\smallskip}
Algorithm & LJ160.1 & LJ160.1S & LJ80.1 & LJ80.1S\\\hline\noalign{\smallskip}
  MIGraph   & $\bf 0.0080$      & $0.0112$      & $\bf{0.0111}$ & $0.0154$\\
  miGraph   & $0.0084$ & $\bf 0.0094$      & $0.0118$      & $\bf{0.0113}$\\
  MI-Kernel & $0.0116$ & $0.0127$      & $0.0174$      & $0.0219$\\\hline\noalign{\smallskip}
  DD        & $0.0852$ & $0.0052$ & N/A      & $0.1116$\\
  BP-MIP    & $0.0398$ & $0.0731$ & $0.0487$ & $0.0752$\\
  RBF-MIP   & $0.0108$ & $0.0075$ & $0.0167$ & $0.0448$\\

\noalign{\smallskip}\hline
\end{tabular}
\end{center}\vspace{-4.5mm}
\end{table}

\section{Conclusion}

Previous studies on multi-instance learning typically treated instances in the bags as i.i.d. samples, neglecting
the fact that instances within a bag are extracted from the same object, and therefore the instances are rarely
i.i.d. intrinsically and the relations among instances may convey important information. In this paper, we
propose two methods which treat the instances in an non-i.i.d. way. Experiments show that our proposed methods
are simple yet effective, with performances highly competitive with the best performing methods on several
multi-instance classification and regression tasks. Note that our methods can also handle i.i.d. samples by using
identity matrix.

An interesting future issue is to design a better graph kernel to capture more useful structure information of
multi-instance bags. Applying graph edit distance or metric learning methods to the graphs corresponding to
multi-instance bags is also worth trying.
The success of our proposed methods also suggests that it is possible to improve other multi-instance learning
methods by incorporating mechanisms to exploit the relations among instances, which opens a promising future
direction. Moreover, it is possible to extend our proposal to other settings such as generalized multi-instance
learning, multi-instance semi-supervised learning, multi-instance active learning, multi-instance multi-label
learning, etc.

\

\section*{Acknowledgments}

This work was supported by NSFC (60635030, 60721002), JiangsuSF (BK2008018) and Jiangsu 333 Program.


\bibliography{miGraph}
\bibliographystyle{plain}

\end{document}